\documentclass{ecai}
\usepackage{times}
\usepackage{graphicx}
\usepackage{latexsym}
\usepackage{enumerate}
\usepackage{hyperref}

\begin{document}

% \title{Guidelines for Preparing a Paper for the \\
% European Conference on Artificial Intelligence}
\title{Little Motion, Big Results: Using Motion Magnification to Reveal Subtle Tremors in Infants}

\author{Girik Malik\institute{Khoury College of Computer Sciences, Northeastern University, 360 Huntington Avenue, Boston, MA 02115, USA, email: gmalik@ccs.neu.edu} $^{,}$\institute{Labrynthe Pvt. Ltd., New Delhi, India} \and Ish K. Gulati \institute{Center for Perinatal Research, Abigail Wexner Research Institute, Nationwide Children’s Hospital, 575 Children’s Crossroads, Columbus, OH 43215, USA, email: ish.gulati@nationwidechildrens.org} $^{,}$\institute{Department of Pediatrics, The Ohio State University College of Medicine, Columbus, OH, USA}  }
% \author{Anonymous Authors \institute{Paper under double blind review}}

\maketitle
\bibliographystyle{ecai}

\begin{abstract}
  Detecting tremors is challenging for both humans and machines. Infants exposed to opioids during pregnancy often show signs and symptoms of withdrawal after birth, which are easy to miss with the human eye. The constellation of clinical features, termed as Neonatal Abstinence Syndrome (NAS), include tremors, seizures, irritability, etc. The current standard of care uses Finnegan Neonatal Abstinence Syndrome Scoring System (FNASS), based on subjective evaluations. Monitoring with FNASS requires highly skilled nursing staff, making continuous monitoring difficult. In this paper we propose an automated tremor detection system using amplified motion signals. We demonstrate its applicability on bedside video of infant exhibiting signs of NAS. Further, we test different modes of deep convolutional network based motion magnification, and identify that dynamic mode works best in the clinical setting, being invariant to common orientational changes. We propose a strategy for discharge and follow up for NAS patients, using motion magnification to supplement the existing protocols. Overall our study suggests methods for bridging the gap in current practices, training and resource utilization.
\end{abstract}

% Keywords:
% Motion magnification
% convolutional network
% deep learning
% computer vision
% Neonatal Abstinence Syndrome
% Discharge strategy

\section{INTRODUCTION}
Infants born to mothers taking prescribed or recreational opioids during pregnancy, often show signs of withdrawal after birth. The constellation of these withdrawal symptoms, known as Neonatal Abstinence Syndrome (NAS), include but are not limited to tremors, seizures, shrieking cry, increased muscle tone and irritability. Seizures are one of the most concerning and life threatening symptoms, which account for 8\% in Methadone users  \cite{methadone}. In the U.S., incidence of NAS has risen six-fold from 2006 to 2016 affecting between 6 and 20 newborns per 1000 live US births \cite{ref2, ref3}.

Unknown probability as well as multitude of symptoms pose a unique challenge to appropriately diagnose NAS when all exposed infants test positive for drug tests on body fluids but not all show troublesome symptoms. A validated scale called Finnegan Neonatal Abstinence Syndrome Scoring System (FNASS) is widely used to monitor and manage therapies \cite{finnegan, ref1}. Opioid exposed infants are initially admitted to a newborn nursery for monitoring and care. Infants may take up to 5 days to metabolize certain drugs taken by the mother before manifesting signs of withdrawal. Infants with qualifying scores for pharmacologic therapy are transferred to a special care nursery. Opioids and their derivatives are the mainstay choices, regardless of the nature of opioid exposure during antenatal period.

However, the absence of a standardized therapy protocol for the treatment of NAS makes FNASS the prime determinant for NAS treatment, which is based on highly subjective evaluation. Most of the primary centers in both rural and urban opioid endemic areas lack trained nurses for FNASS scoring, and as a result, infants are transferred to a higher center for optimal scoring, monitoring and treatment. About one-half of the infants are born at resource limited hospitals, and need to be transferred to tertiary care centres for optimal management \cite{ref4}. 

We aim to overcome this subjectivity and limitation of highly skilled nursing training using vision-based objective monitoring and evaluation technique. Our hypothesis is based on the principle of objective monitoring evaluation of tremors, mitigating the need for trained nurses, minimising nursing exposure and allowing the possibility of remote monitoring. The goal is to capture tremor objectively in an affected infant and supplement it with other parameters in the scale. We use motion magnification \cite{wu} to amplify tremors, which are constant, involuntary, spontaneous, and repetitive movements at high frequency but low amplitude, and are commonly confused with common newborn jitters and other newborn movements.

% \noindent \textbf{Contributions}:
\vspace{2mm}
\noindent The main contributions of this paper are as follows:
\vspace{-2mm}
\begin{enumerate}[-]
    \item System for continuous monitoring of NAS patients using Motion Magnification
    \item Converting subjective visual evaluations of NAS patients to objective evaluations
    \item Proposal of discharge strategy for NAS patients
\end{enumerate}

% The paper is organized as:
We start with background on Motion Magnification in Section \ref{sec:motionmag} and Neonatal  Abstinence  Syndrome (NAS) in Section \ref{sec:nas}. We describe the specifics of an automated tremor detection system in Section \ref{sec:automatedtremor}, starting with the details of the network used and experiments in Section \ref{sec:methods}, and results in Section \ref{sec:results}. We propose a follow-up and discharge strategy for patients in Section \ref{sec:followup}. The challenges, strengths and limitations of this work are discussed in Section \ref{sec:discussion}. The envisioned future direction of the current research and its applicability to other domains is discussed in Section \ref{sec:conclusionandfuture}.

\section{BACKGROUND}
\subsection{Motion Magnification}
\label{sec:motionmag}
Motion magnification can be widely classified into two categories, Lagrangian and Eulerian. In this paper, we use the Eulerian approach \cite{wu}, which decomposes video frames into representations useful for manipulating motion, without explicitly tracking the target in every frame. 

Mathematically, let $I(x,t)$ denote the image intensity at position $x$ and time $t$. For translational motion(s), we can express the observed intensities with respect to a displacement function $\delta(t)$, such that $I(x,t) = f(x+\delta(t))$, while the reference frame is given by $I(x,0)=f(x)$. The goal of motion magnification is to produce a magnified image representation $\hat I$, such that
$$\hat I(x,t)=f(x+(1+\alpha)(\delta (t)))$$
for some amplification factor $\alpha$.

For this work, we used a fully convolutional encoder-manipulator-decoder network, as described in \cite{oh}. The network learns and applies filters directly to the examples, instead of using temporal filters. However, the learned representations can be extended for use with temporal filters for frequency-based motion selection. There are two main modes considered for this work, static and dynamic. In case of static amplification, the first frame is used as a reference, i.e. $(X_0, X_t)$ frames are used as input; whereas dynamic amplification uses the previous frame as reference, i.e. $(X_{t-1} ,X_t)$ are used as input, magnifying the difference between consecutive frames. We also talk about using temporal filters \cite{wadhwa}, please see Section \ref{sec:methods}.

\subsection{Neonatal Abstinence Syndrome (NAS)}
\label{sec:nas}
NAS represents a clinical phenotype, as a result of opioid exposure during the antenatal period. Opioids can easily cross the fetal blood brain barrier, accumulate in the fetus leading to prolonged half life, thereby increasing the severity of withdrawal symptoms after birth \cite{opioidbloodbrain}. A persistent exposure to high dosage of opioids during pregnancy results in increased stimulation of neurotransmitters \cite{opioidstimulation}. Noradrenaline is the most sensitive neurotransmitter in opioid withdrawal and is secreted from Locus coeruleus of the fetal brain \cite{little}. Tremor is a known symptom of a hypernoradrenergic state \cite{tremorhyper}. 
 
The displacement caused by tremors is an important factor in classifying NAS patients. While sometimes imperceptible to the naked eye, these movements can be identified by amplification of motion using techniques like motion magnification \cite{wu, freeman}. In case of NAS patients, there are observed sudden, non-purposeful, and non-repetitive movements as well, causing major displacement of limbs. The distinction of these voluntary movements from the involuntary ones is fairly subjective in nature, making the quantitative objectification a challenging problem. 
We would also like to highlight the dearth of datasets in the direction of objective evaluation of infants with NAS, and video datasets for tremors, making it a nascent field. 

\section{AN AUTOMATED TREMOR DETECTION SYSTEM}
\label{sec:automatedtremor}
\subsection{Experiments}
\label{sec:methods}
Our study applies the neural network from \cite{oh} on an open-source bedside video of a baby exhibiting signs of NAS. For control, we used the video of a sleeping baby from Wu et al. \cite{wu}.

We use the deep convolutional neural network described in Oh et al. \cite{oh}, with three primary components, namely, spatial decomposition filters, representation manipulator, and reconstruction filters, which are designed as encoder, manipulator and decoder networks. The encoder and decoder networks are fully convolutional and use residual blocks for generating high-quality images. Additionally, the encoder and decoder also downsample and upsample the input using strided convolution and nearest-neighbour upsampling respectively. The manipulator works by multiplying the difference between the two representations found by the encoder, based on the given amplification factor (Please see \cite{oh} for details).

\begin{figure}
  \centering \includegraphics[width=0.5\textwidth]{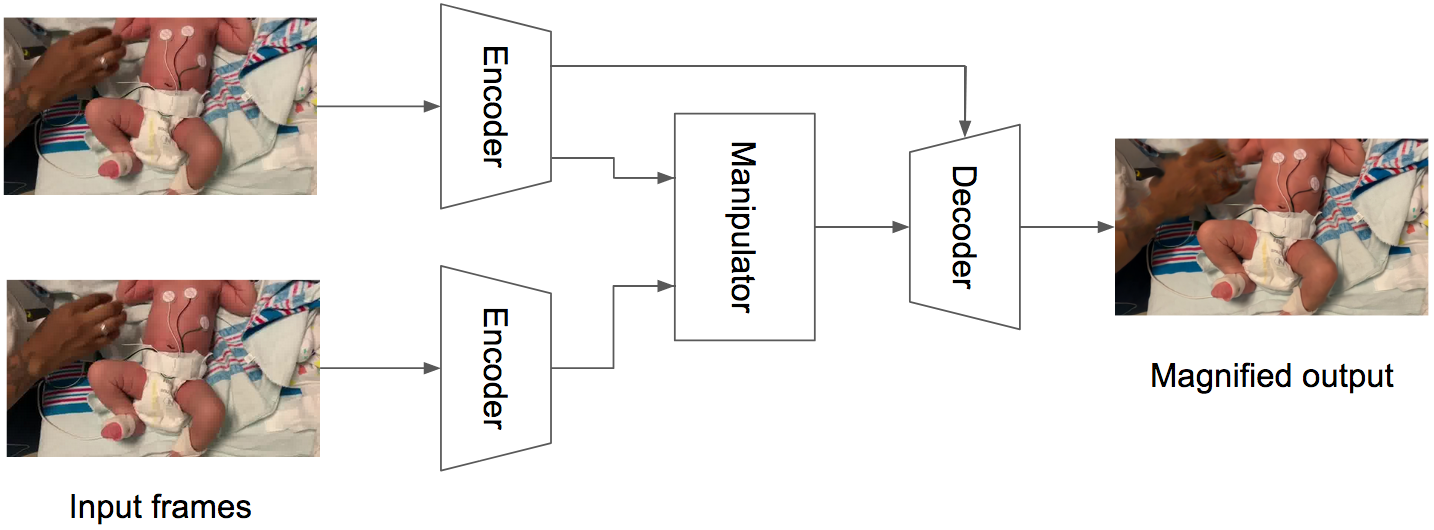}
  \vspace{-5mm}
  \caption{Schematic of motion magnification applied using the described architecture. Two adjacent frames are given as input to the fully convolutional encoder network for extracting shape and texture representations. These representations are further fed to a manipulator network, for amplifying the motion signals. The manipulated representation is then fed to a decoder network that upsamples the representation to construct the motion-amplified frames.}
  \label{fig:1}
\end{figure}

Two frames from the video are given as input to the encoder network. In case of dynamic mode, the frames are adjacent, while in case of static mode, the input is first frame and the one at time $t$. The encoder behaves like a spatial decomposition filter that extracts the shape representations from each image separately. The representation is then fed to the manipulator for amplifying the motion. Finally, the amplified representation is fed to the decoder, which reconstructs the modified representation into an individual magnified frame. See Fig \ref{fig:1}.  

In addition to static and dynamic mode, we also show the application of linear temporal filters, which have worked well in case of linear shape representations \cite{tf1, tf2, tf3}. Using the shape representation, extracted from the encoder network, the difference operation in the manipulator network is replaced by a pixel-wise temporal filter across the temporal axis. This new, temporally-filtered shape representation is fed to the decoder network for generating magnified frames.

We used weights from the network pre-trained on the synthetic dataset from \cite{oh}. The network is trained using $\ell_1$-loss and ADAM Optimizer \cite{adam}, with a learning rate of $10^{-4}$ and no weight decay. The dataset consists of background images from MS COCO dataset \cite{coco}, superposed on objects from PASCAL VOC dataset \cite{pascal}. We tested the network in static and dynamic modes using $\alpha = 10$, while for temporal mode, we set $\alpha = 20$. %, as tabulated in Table \ref{tab:1}. 

\vspace{-5mm}
% \begin{table}
% \caption{\label{tab:1} Amplification Factor ($\alpha$) used for different modes}
% \begin{center}
% \vspace{-5mm}
% \begin{tabular}{|l|l|}
% \hline
% \textbf{Mode    }   & \textbf{Amplification Factor} \\ \hline
% Static          & 10                     \\ \hline
% Dynamic         & 10                     \\ \hline
% Temporal Filter     & 20                     \\ \hline
% \end{tabular}
% \end{center}
% \end{table}
% \vspace{-5mm}

\subsection{Results}
\label{sec:results}

\begin{figure*}
  \centering \includegraphics[width=0.9\textwidth]{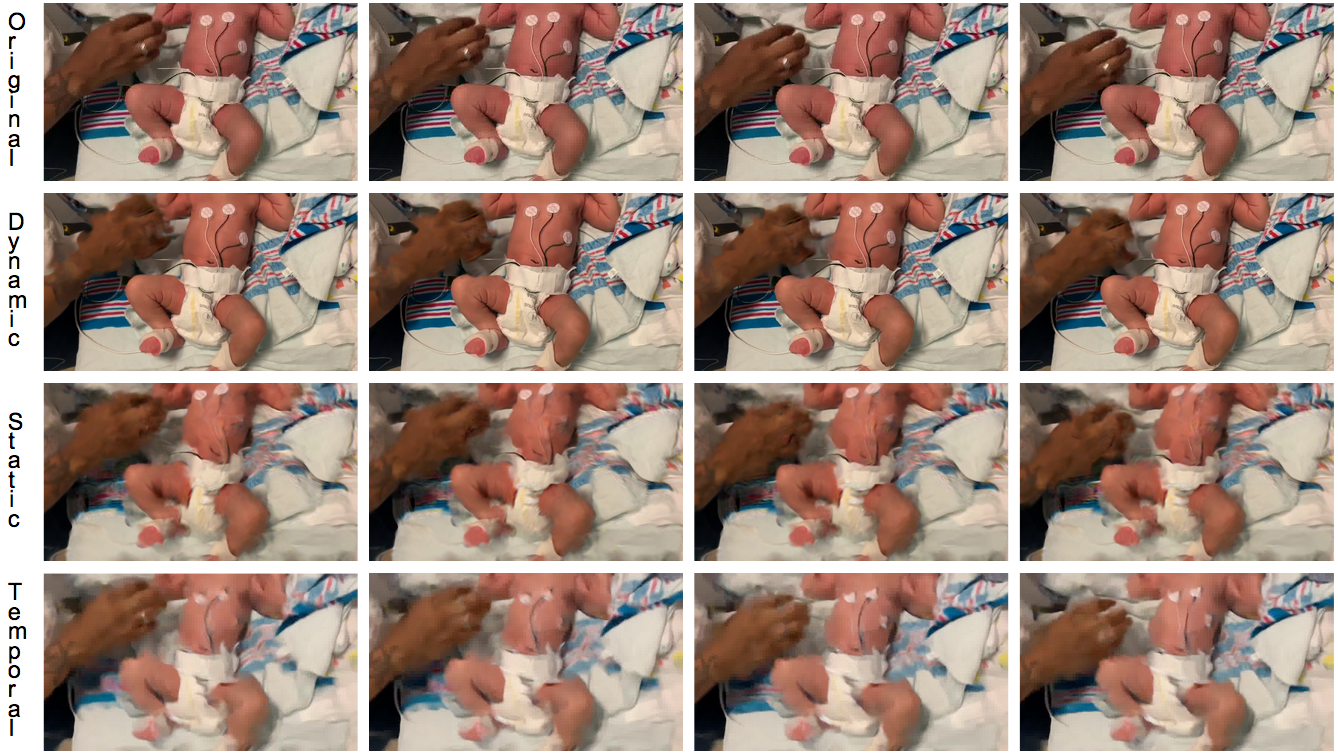}
  \caption{Result from dynamic, static and temporal filter mode of amplification. The top row shows frames from the original video, while the subsequent rows show the corresponding motion-magnified frames. Observe how the subtle motion in the infant’s body is picked up and amplified by the network, while the voluntary movement of caregiver’s hand is distorted as an artefact. The stationary objects are not influenced by the dynamic mode.}
  \label{fig:2}
\end{figure*}

We demonstrate the application of static, dynamic and temporal filter \cite{wadhwa} based magnification approaches, to a bedside video of an infant exhibiting the signs of NAS. We compare the approach with application of the same algorithms to a sample baby video, as used in \cite{wu}.

Our results clearly indicate that the dynamic method, magnifying the difference between consecutive frames, has fewer edge artefacts compared to static and temporal mode. For regular actions, like breathing, the difference in the original and magnified video is insignificant. During tremors, the video processed using dynamic mode, starts exhibiting magnified movements, wherein the body moves in a subtle pattern, while the limbs seem to move in a more hysterical and uncontrollable manner. The caregiver’s hand in the scene is also distorted in the magnified frame, and not amplified. Dynamic mode is also invariant to orientational changes during the video. 

In static mode, with the first frame taken as reference, body movement is less magnified, compared to the surroundings. Keeping the first frame as anchor, it can magnify the objects with limited displacement from their original position across the frames. It suffers from ringing artefacts and limits the ability to operate in conditions with frequent orientational changes (rotations). The temporal filter mode also suffers from edge artefacts, given its inability to learn complex limb movements with the linear temporal filters. Stationary objects are not amplified, but seem to be distorted in the static and temporal filter case, as possible edge artefacts due to the distorted motion of infant’s limbs. Results comparing the original and magnified frames are shown in Fig. \ref{fig:2}.

\section{FOLLOW UP AND DISCHARGE STRATEGY FOR NAS PATIENTS}
\label{sec:followup}
NAS infants often need to follow up for rebound symptoms using Finnegan scoring for upto 2 to 5 days after therapy is discontinued. In borderline results, infants may be kept in hospital for longer periods \cite{nasfollowup}. This technology may have applications for improving discharge protocols in such situations. In the current and post COVID era, focus will be on  minimizing the number of patients in the hospital,  shortening the length of stay and accessible remote monitoring. Such technology could help with monitoring infants at home because of its low cost and ease of operability. It is possible to bring the cost of setup inline with the other at-home monitoring equipments using low-resource hardware, and to bring down the computational costs by using networks like MobileNets \cite{mobilenets} for network backbone. 

\section{DISCUSSION}
\label{sec:discussion}
Neonatal abstinence syndrome (NAS) management has unintended troublesome consequences including logistical challenges of infant transfer, mother-infant dyad separation, lack of kangaroo care of the separated infant and prolonged hospital stay, stretching resources. Socio-economic disparity has been reported in allocation of resources for optimal management \cite{ref4}. In the current and post COVID era, we expect the healthcare system to face deeper challenges. Some low resource models are already struggling. Those struggling earlier, face imminent closures. One way of emerging successfully from this crisis is to integrate current technology in our healthcare practice, not only with the medical devices, but also bringing solutions for training, objectification of clinical subjectivity, remote monitoring, and generating and utilizing the data to improvise. 

In this paper, we have addressed one major issue of non standardized clinical monitoring. The current standard-of-care for patients with NAS is still dependent on subjective evaluations which are prone to human errors \cite{naserrors}. While it is impossible to argue for a complete automation of anything in healthcare, there are certain areas that need innovation to be at par with standardization in other domains. In this paper, we make the first step towards such a standardization, by objectifying a largely subjective Finnegan scoring for NAS patients. Our use of motion magnification as a tool to detect and amplify tremors in infants that are imperceptible to naked eyes could help in better continuous monitoring of patients, that otherwise requires highly skilled nurse practitioners monitoring in intermittent intervals. The current discharge and followup strategy for patients with NAS is also very loosely defined, without an objective way of catering to misclassifications.

For our pilot study, we tested three different modes of motion magnification, and found that dynamic mode performed best with the current video. Our observations for static mode were coherent with the expected behaviour for the current video, given the use of the first frame as reference. The temporal filter mode seems to produce edge artefacts, and needs more analysis with domain specific data and better kernels to select small motions of interest. We believe the currently implemented linear temporal filter might not be suitable to learn the representations of complex non-linear motion. We propose a video camera monitoring the infant with a monocular video stream of 640x480 at 30-45 frames per second, fixed to the bedside. This setup gives a continuous video stream, which is processed using Eulerian Video Magnification \cite{wu, oh}

It is often easy to confuse tremors with tremor mimickers at the bedside. Physical manifestations of tremor in NAS infants may look like myoclonus (sudden jerking), jitteriness or fine tremors, and are often misinterpreted as epileptic seizures, requiring electroencephalogram (EEG) \cite{palla}. Motion magnification will be capable of diagnosing and aiding clinical diagnosis of seizures with EEG. We propose that once a tremor signature of NAS is established, clinical seizures of NAS will help correlate EEG findings of epileptic focus. We hope further research in this area will explore more opportunities for characterisation of NAS tremors, and allow healthcare practitioners to recommend personalised therapy and management plans.

\textbf{Limitations}: The small size and type of currently available datasets makes it challenging to be used with deep learning methods. There is a need for more extensive data collection and its standardized protocols approved by Institutional Review Board(s) (IRB), specifically videos of tremors and seizures, for vision related methods, to train humans and machines alike. In our study, we were limited by the dataset size for the same reason. 

\textbf{Strengths}: The video of infant with NAS used has rotational changes of about 90 degrees in the latter half (not shown in result images), where the dynamic mode performed as well as in the earlier part, showing its robustness to orientational changes. The study presented is first to report how to objectively capture NAS tremor using motion magnification. The possible use of low-resource hardware allows easy scale-up of the system for monitoring patients at home and in remote areas. However, we need a clinical feasibility and validation study, along with a diverse video dataset, to compare this innovative technology with standard of care. 

\section{CONCLUSION AND FUTURE DIRECTIONS}
\label{sec:conclusionandfuture}
We have shown how the very subtle motion of the tremor in the center of the infant’s body is picked up by the motion magnification network, while the voluntary movement of the care-giver’s hand is distorted, and not amplified. Additionally, we highlighted the problems in the existing subjective evaluations, and made proposals of bridging those gaps with innovative techniques using deep neural networks. This project aligns with American Academy of Pediatrics goals in addressing both its key issues of health disparities and health equities by empowerment of low resource centers in disproportionately higher prevalence of opioid addicted mothers and infants with NAS. We make some suggestions based on important observations in the field, that we believe will improve monitoring of NAS patients, and help with better infant care in general.

Infants with NAS have a more shrieking high pitched cry recorded as a characteristic acoustic signature versus low pitched cry of healthy infants. As next steps, we are investigating differences in acoustics for detecting high pitched cry, and if they can be combined with our vision-based model to add more sensitivity to the sample. Once validated, an automated video based motion magnification tool can be used to train care providers to understand the mechanism of these pathophysiological manifestations, in low-resource settings. Further, we propose to formulate this setup to a scoring tool for patients showing unique tremor signatures during and after the treatment of NAS, to strengthen the existing protocols. We also envision to extrapolate automatic tremor detection to monitor patients with stroke and Parkinson’s disease in nursing homes.

\ack
We would like to thank the referees for their comments and suggestions, which helped improve this paper considerably. GM would like to thank his advisor Prof. Ennio Mingolla, Northeastern University, for letting him work on this research, which is not directly related to his PhD work. IKG would like to thank Dr. Deepak Gulati, Vascular Neurologist, The Ohio State University College of Medicine for his helpful insights.
\bibliography{ecai}

\end{document}